\documentclass[10pt,twocolumn,letterpaper]{article}

\usepackage{cvpr}              

\definecolor{cvprblue}{rgb}{0.21,0.49,0.74}
\usepackage[pagebackref,breaklinks,colorlinks,allcolors=cvprblue]{hyperref}
\usepackage{subcaption}
\usepackage{amsmath}
\usepackage{graphicx} 

\usepackage{multirow}
\usepackage{vector}
\usepackage{booktabs}
\usepackage{pgfplots}
\pgfplotsset{compat=1.18}
\usepackage{siunitx}
\usepackage{algorithm}
\usepackage[noend]{algcompatible}
\usepackage{hyperref}

\usetikzlibrary{arrows}
\usetikzlibrary{angles}

\def\paperID{*****} 
\def\confName{CVPR}
\def\confYear{2025}

\title{Beyond Gradient Averaging in Parallel Optimization:\\ Improved Robustness through Gradient Agreement Filtering}

\author{Francois Chaubard\\
{\tt\small fchaubar@stanford.edu}
\and
Duncan Eddy\\
{\tt\small deddy@stanford.edu} \\
Stanford University\\
\and
Mykel J. Kochenderfer\\
{\tt\small mykel@stanford.edu}
}

\begin{document}
\maketitle
\begin{abstract}

We introduce Gradient Agreement Filtering (GAF) to improve on gradient averaging in distributed deep learning optimization. Traditional distributed data-parallel stochastic gradient descent involves averaging gradients of microbatches to calculate a macrobatch gradient that is then used to update model parameters. We find that gradients across microbatches are often orthogonal or negatively correlated, especially in late stages of training, which leads to memorization of the training set, reducing generalization. In this paper, we introduce a simple, computationally effective way to reduce gradient variance by computing the cosine distance between micro-gradients during training and filtering out conflicting updates prior to averaging. We improve validation accuracy with significantly smaller microbatch sizes. We also show this reduces memorizing noisy labels. We demonstrate the effectiveness of this technique on standard image classification benchmarks including CIFAR-100 and CIFAR-100N-Fine. We show this technique consistently outperforms validation accuracy, in some cases by up to 18.2\% compared to traditional training approaches while reducing the computation required nearly an order of magnitude because we can now rely on smaller microbatch sizes without destabilizing training. We release our code for reproducibility and easy adoption at: \url{https://github.com/Fchaubard/gradient_agreement_filtering} 

\end{abstract}    
\section{Introduction}
\label{sec:intro}

With increasingly large deep learning models, hardware constraints often limit the feasible batch size that can fit within the memory (VRAM) of even the most powerful GPUs. Consequently, machine learning practitioners must distribute the training across hundreds or thousands of GPUs using Distributed Data Parallelism (DDP), Distributed Model Parallelism (DMP), or a combination of both \cite{li2020pytorch,dean2012large,shoeybi2019megatron,you2020large,hoffmann2022training}. As a result, the traditional minibatch in stochastic gradient descent (SGD) has been replaced with microbatches and macrobatches \cite{Piao_2023}. A microbatch is defined as the samples processed by a single forward and backward pass to produce a microbatch gradient, often called a micro-gradient. Microbatches are typically produced on a per GPU basis and then shared across all of the other GPUs to calculate the macrobatch. A macrobatch is the union of all microbatches. Each micro-gradient is summed and then normalized to compute the macro-gradient, which is then used to update the model. In practice, the microbatch size is chosen to maximize the VRAM utilization on a single GPU or computation node. During the aggregation of micro-gradients, practitioners leverage the Ring-AllReduce algorithm \cite{baidu2017ringallreduce} to efficiently aggregate micro-gradients across all computation nodes. Ring-AllReduce relies on sequential summation and normalization to ensure synchronization of gradient values across all nodes without each node needing to retain multiple micro-gradient copies in memory. Once all gradients from the macrobatch have been aggregated a parameter update is performed and the process is repeated.

\begin{figure*}[t!]
  \centering
  \begin{subfigure}{0.495\linewidth}
    \includegraphics[width=\linewidth,  trim=0 0 0 17mm, clip]{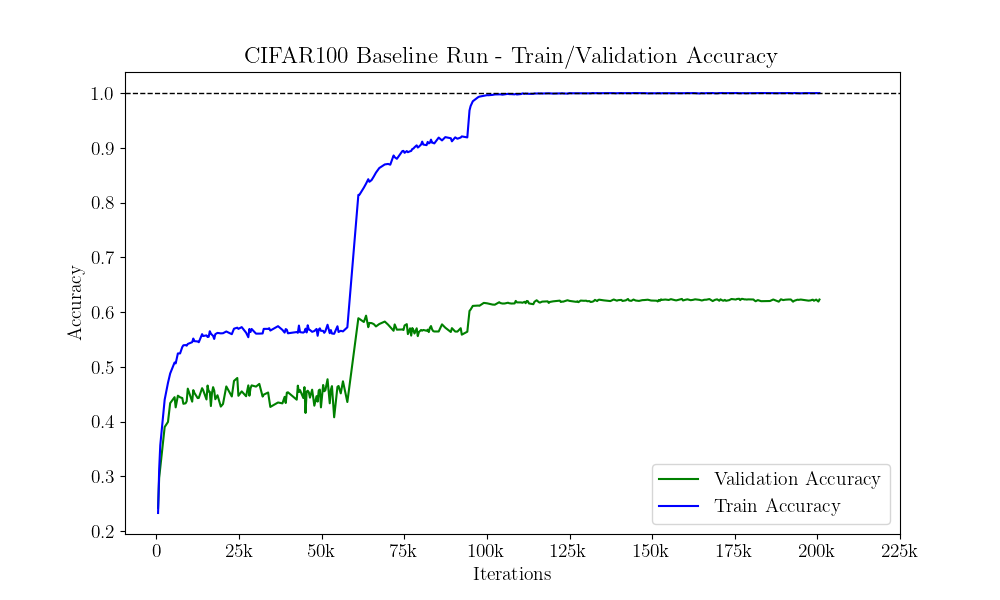}
    \caption{Train and validation accuracy over training on CIFAR-100}
    \label{fig:trainingval_cifar100}
  \end{subfigure}
  \hfill
  \begin{subfigure}{0.495\linewidth}
    \includegraphics[width=\linewidth,  trim=0 0 0 17mm, clip]{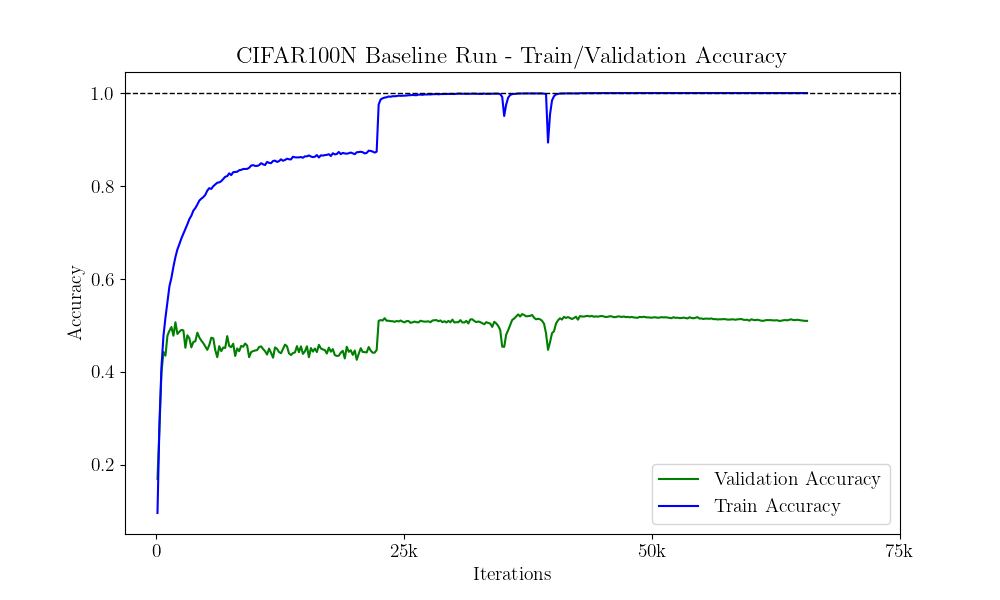}
    \caption{Train and validation accuracy over training on CIFAR-100N-Fine}
    \label{fig:trainingval_cifar100n}
  \end{subfigure} \\
  \begin{subfigure}{0.495\linewidth}
    \includegraphics[width=\linewidth,  trim=0 0 0 17mm, clip]{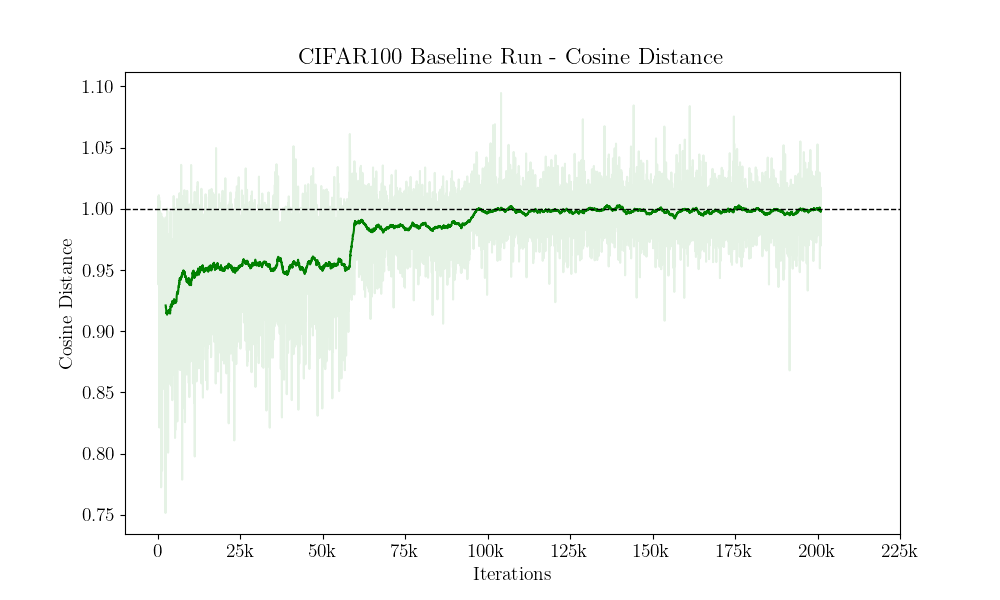}
    \caption{Cosine distance over training on CIFAR-100}
    \label{fig:cosdist_cifar100}
  \end{subfigure}
  \hfill
  \begin{subfigure}{0.495\linewidth}
    \includegraphics[width=\linewidth,  trim=0 0 0 17mm, clip]{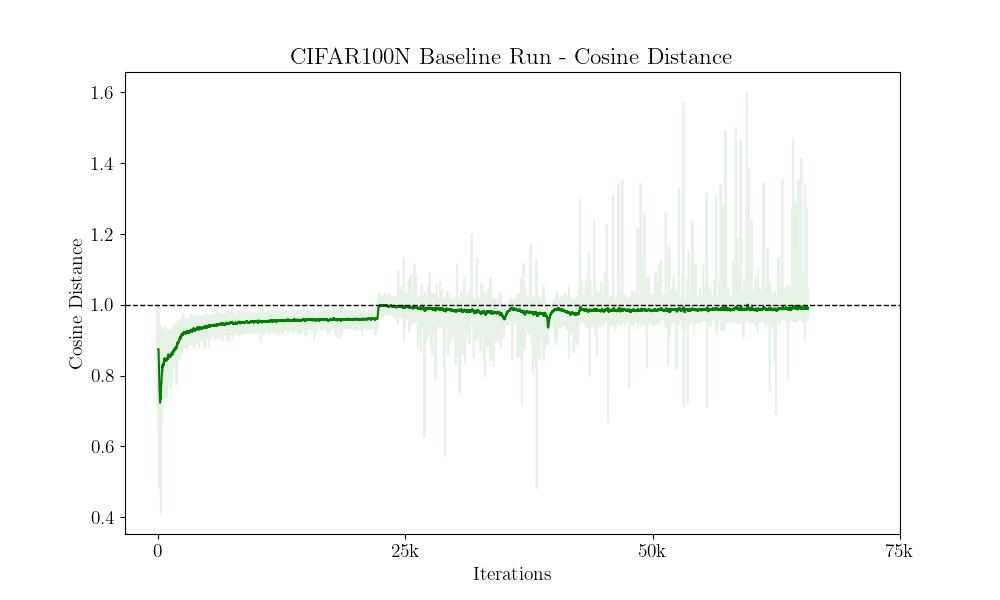}
    \caption{Cosine distance over training on CIFAR-100N-Fine}
    \label{fig:costdist_cifar100n}
  \end{subfigure}
  \caption{We train a ResNet18 on CIFAR-100 (left) and CIFAR-100N-Fine (right). We show train and validation accuracy  over iterations (top). The rolling average of cosine distances is shown in dark green and the raw cosine distances during training in light green (bottom). In late stages of training in all runs, as training accuracy plateaus, the cosine distance between micro-gradients approaches 1 with many micro-gradients diverging even further up to 1.1 for CIFAR-100 and 1.6 for CIFAR-100N-Fine. }
  \label{fig:val_cosd_cifar}
\end{figure*}

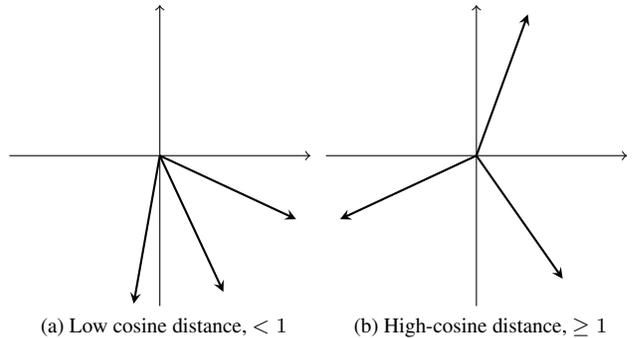
\begin{figure}[ht!]
  \centering
  \begin{subfigure}{0.495\linewidth}
    \begin{tikzpicture}
        \draw[->] (-2, 0) -- (2, 0) node[right] {};
        \draw[->] (0, -2) -- (0, 2) node[above] {};
    
        
        \foreach \angle in {260, 295, 335} { 
            \draw[->, thick, >=stealth] (0, 0) -- ({2*cos(\angle)}, {2*sin(\angle)})
                node[shift={(0.2,0.2)}, anchor=south] {};
        }
    \end{tikzpicture}
    \caption{Low cosine distance, $<1$}
    \label{fig:micrograd_aligned}
  \end{subfigure}
  \hfill
  \begin{subfigure}{0.495\linewidth}
    \begin{tikzpicture}
        \draw[->] (-2, 0) -- (2, 0) node[right] {};
        \draw[->] (0, -2) -- (0, 2) node[above] {};
    
        
        \foreach \angle in {70, 205, 305} {
            \draw[->, thick, >=stealth] (0, 0) -- ({2*cos(\angle)}, {2*sin(\angle)})
                node[shift={(0.2,0.2)}, anchor=south] {};
        }
    \end{tikzpicture}
    \caption{High-cosine distance, $\geq 1$}
    \label{fig:micrograd_misaligned}
  \end{subfigure}
  \caption{Visualization of cosine distance between micro-gradient batches in 2D. Aligned gradients have low cosine distance (left), while orthogonal or negatively correlated gradients have high cosine distance (right).}
  \label{fig:micrograd_visual}
\end{figure}

However, there is an underexplored question: is averaging all micro-gradients the best thing to do all of the time? Furthermore, are micro-gradients ever orthogonal or, worse, negatively correlated with each other during training? If so, what does this imply? This has been recently explored in the context of reinforcement learning for both multi-constraint  \cite{yao2023gradientshapingmulticonstraintsafe} and multi-task optimization \cite{yu2020gradientsurgerymultitasklearning} where gradients with respect to specific constraints or tasks are compared using cosine distance and update a projected component of the conflicting gradient onto the update direction or skip the gradient update altogether if the direction violates constraints. However, this has yet to be developed as a general optimization procedure, specifically in the context of distributed training. The first question we explore is what happens during typical training. Are the micro-gradients always correlated or are they orthogonal or divergent? To measure this, we compute the cosine distance between micro-gradients before averaging them over the course of training ResNet18 \cite{he2016deep} on both CIFAR-100 \cite{krizhevsky2009learning} and CIFAR-100N-Fine \cite{wei2022learning}. The cosine distance $D_c(\mathbf{x},\mathbf{y}) \in [0, 2]$ between two vectors $\mathbf{x}$ and $\mathbf{y}$ is used as a measure of divergence between the vectors. A cosine distance of 0 means two vectors are perfectly correlated, a cosine distance of 1 means they are orthogonal, and a cosine distance of 2 means they perfectly negatively correlated (pointed in the opposite directions). We find that during training micro-gradients are often orthogonal or negatively correlated with each other resulting in cosine distances close to or above 1 especially in late stages of training, as shown in \Cref{fig:cosdist_cifar100} and \Cref{fig:costdist_cifar100n}. The observation that micro-gradients are significantly misaligned, to the point of orthogonality or beyond, suggests that each microbatch offers a meaningfully different approximation of the true loss surface. This indicates that the optimization procedure should be cautious of stepping with these gradients as there is no consensus on which direction to take. This can be intuitively visualized in two dimensions as seen in \Cref{fig:micrograd_visual}.

\begin{figure}[ht]
    \centering
    \includegraphics[width=0.975\linewidth]{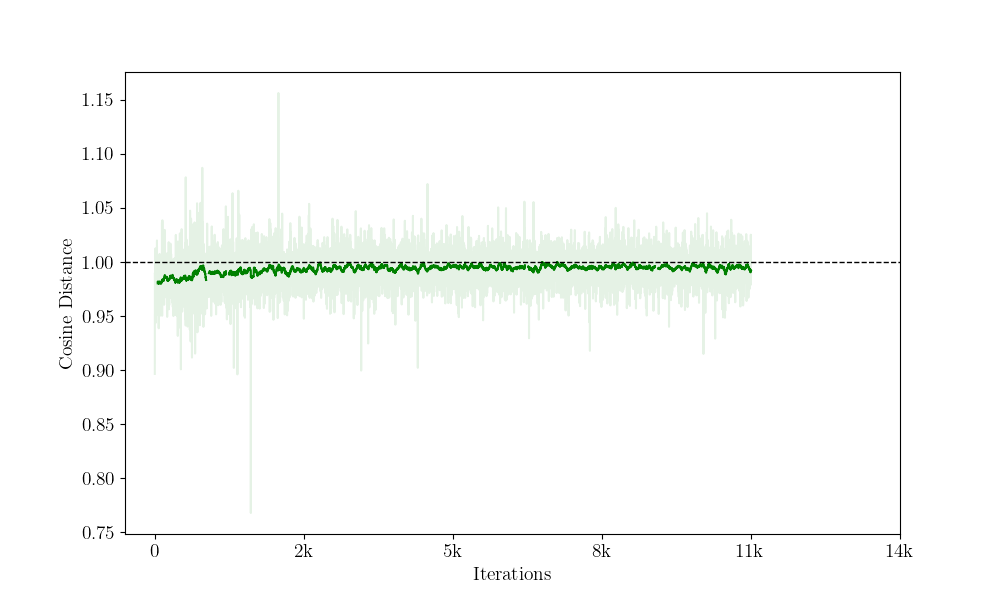}
    \caption{Cosine distances between micro-gradients during the later stages of a baseline ViT-L/16 run on ILSVRC12}
    \label{fig:baseline_vit_run}
\end{figure}

When training image classification models, we find that this pattern exists across all tested model sizes and datasets. Micro-gradients exhibit a high cosine distance (close to or above 1) during both very early and late stages of training. This was observed in both smaller models such as ResNet18 (11 million parameters) on CIFAR-100, as shown in \Cref{fig:cosdist_cifar100} and \Cref{fig:costdist_cifar100n}, and  larger models such as ViT-L/16 (300 million parameters) \cite{dosovitskiy2021an} trained on ImageNet (ILSVRC12) \cite{5206848}, as shown in \Cref{fig:baseline_vit_run}. Large cosine distances in both early and late stages of training can be explained by the information bottleneck principle \cite{tishby2015deep}. In early stages of training, the randomly initialized weights lack well-formed kernels that are able to produce activations with high mutual information with the training set, and each microbatch may easily disagree on where in the model to place each of these kernels or features. In later stages of training, when these kernels or features are well-formed and training accuracy nears 100\%, micro-gradient misalignment implies that at least one of the micro-gradient directions is a step into memorizing each individual microbatch, instead of a step into further generalizing on the validation set. In either case, we show skipping these steps results in more stable training and less overfitting which is the core idea of this paper.

\begin{figure}[ht]
    \centering
    \includegraphics[width=0.975\linewidth]{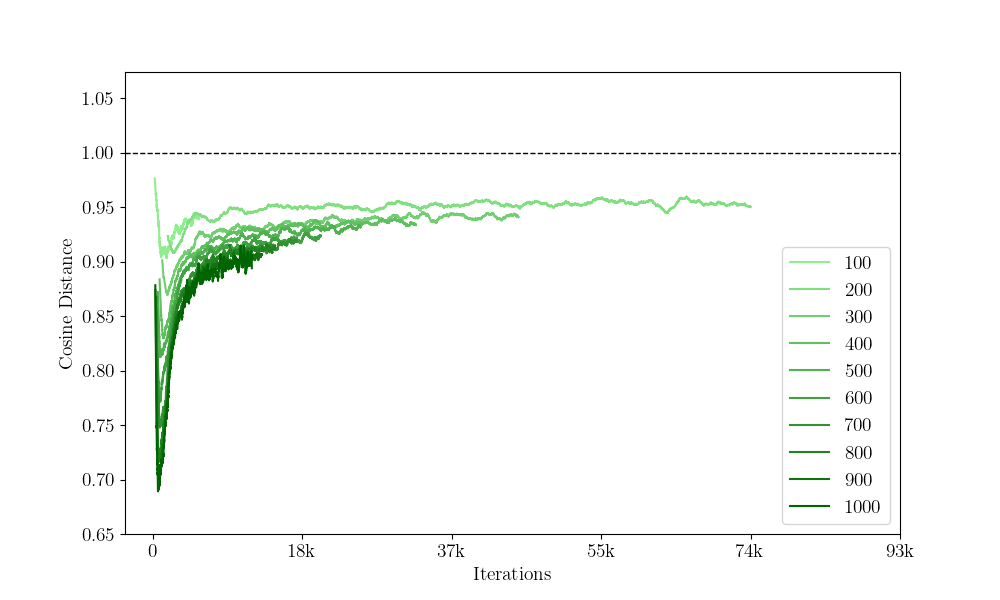}
    \caption{Rolling average of cosine distances between micro-gradients during 10 baseline runs on CIFAR-100 varying batch sizes from 100 to 1000. As the microbatch size increases, the micro-gradients become more and more correlated throughout training.}
    \label{fig:baseline_cifar100_batchsize}
\end{figure}

The conventional way to train deep models in a stable way has been to increase batch size to average out noise-induced variances up to the point of diminishing returns where larger batch sizes beyond some critical point result in a reduction in generalization \cite{keskar2016large,arpit2017closer,mccandlish2018empirical}, with increased computational cost and risk of memorizing noisy labels. We see that as the microbatch size increases the micro-gradients across 2 microbatches become increasingly correlated, as shown in \Cref{fig:baseline_cifar100_batchsize}. This would be true even in the presence of noise in the dataset as per the law of large numbers. When taken to the limit of setting the microbatch size equal to the entire training set, the cosine distance clearly would approach 0. This implies that there is an optimal batch size for every training run. This was explored by \citeauthor{mccandlish2018empirical} where beyond this critical point, called the critical batch size, increasing macro-batch size further only increases computational cost without significant training benefit. However, these critical batch sizes are still often quite large. For example, the current state-of-the-art image classification model uses a batch size of 4096 which requires 8.2TB of VRAM in distributed training. What if we can achieve equal or better generalization at batch sizes far beneath \citeauthor{mccandlish2018empirical}'s critical batch size by leveraging our knowledge of divergent micro-gradients? 

This paper presents a solution that achieves equal to or higher generalization with much smaller batch sizes by calculating the cosine distances between micro-gradients and filtering out those that have high cosine distance from each other during aggregation, prior to performing the gradient update. This approach, called Gradient Agreement Filtering (GAF), leads to fewer accepted micro-gradients which significantly increases generalization and robustness, especially in the presence of noisy data. In \Cref{sec:prior_work} we discuss relevant related works. Next in \Cref{sec:methods} we introduce the core concepts and algorithmic implementation. Finally, in \Cref{sec:experiments} we demonstrate the efficacy of GAF on training ResNet18 and ResNet34 image classifiers on CIFAR-100 and CIFAR-100N-Fine, datasets respectively, where we find that adding GAF improves validation accuracy from 0.2\% to 18\%, with only a batch size of 200 (a microbatch size of 100 and a macrobatch size of 2). This outperforms the non-GAF baseline runs over all batch sizes, up to and including batches as large as 1100. We also show that as microbatch size increases in GAF-based training, this improvement decreases as the signal washes out. This suggests smaller micro-gradients with GAF are better for generalization, robustness to noisy labels, and compute.


\section{Previous Works}
\label{sec:prior_work}

Gradient descent has been fundamental to machine learning since the 1950s. Initial works introduced the basic iterative framework for gradient descent and soon led to the development of stochastic gradient descent (SGD)~\cite{robbins1951sgd}, which computes gradients on small, random subsets of data rather than the entire dataset, enabling efficient optimization in high-dimensional settings. Building upon this foundation, researchers have developed various enhancements to SGD, including \textit{momentum}~\cite{polyak1964some}, which introduces a velocity term to accelerate convergence in regions with low curvature, and \textit{Adam}~\cite{kingma2014adam}, which combines momentum with adaptive learning rates to better handle sparse gradients and noisy data. A recent refinement is, \textit{AdamW}~\cite{loshchilov2017decoupled}, decouples weight decay from the gradient update process, yielding better generalization properties by stabilizing the learning dynamics. In recent years, the deep learning community has produced a substantial body of research that addresses the practical challenges of gradient-based optimization.


To handle the immense computational demands of training large models, researchers have explored various distributed and parallel training frameworks based on the concepts of data and model parallelism. These approaches enable practitioners to scale training across multiple GPUs or compute nodes, facilitating larger batch sizes and reducing the time required for model convergence. Techniques like \textit{Ring-AllReduce}~\cite{baidu2017ringallreduce} allow for efficient gradient aggregation across GPUs, minimizing communication overhead, and memory, enabling synchronous training on high-performance systems. Additionally, asynchronous gradient sharing strategies and parameter servers~\cite{dean2012large} have been proposed to further enhance scalability, though come at the cost of potential staleness in parameter updates.


Adaptive optimization algorithms, including \textit{RMSProp}~\cite{tieleman2012lecture} and \textit{Adam}~\cite{kingma2014adam}, address the limitations of standard SGD by dynamically adjusting learning rates based on historical gradient information. These methods have proven especially useful in handling noisy or sparse gradients, which are common in large-scale deep learning models. Recent advancements, such as layer-wise adaptive moments (\textit{LAMB})~\cite{you2020large} and \textit{AdaBelief}~\cite{zhuang2020adabelief} focus on improving generalization by adapting learning rates according to layer-specific characteristics or reducing reliance on gradient magnitudes to mitigate training instability.


A challenge in deep learning is the balance between fitting the training data well while simultaneously avoiding overfitting and memorizing train noise. Researchers have proposed various strategies to control overfitting, such as data augmentation, dropout~\cite{srivastava2014dropout}, and early stopping~\cite{prechelt1998early}. Recent work on sharpness-aware minimization (\textit{SAM})~\cite{foret2020sharpness} explicitly targets solutions within flatter regions of the loss landscape to promote generalization, which has shown significant promise across various deep learning benchmarks.


Training deep models in the presence of noisy labels is a challenging problem, as noise can lead to memorization of incorrect labels and hinder generalization. Several methods have been proposed to address label noise, including learning from noisy labels~\cite{li2020learning}, co-teaching~\cite{han2018coteaching}, and learning to learn from noisy labels~\cite{ren2018learning}. These methods often rely on a dual-network architecture, where one network acts as a teacher or peer model to guide the student model in selectively learning from clean samples. This approach, however, is computationally expensive as it requires training two instances of the same model in parallel, which scales poorly for large models and datasets. More recent approaches, such as self-supervised over-parametrization (\textit{SOP})~\cite{liu2022self}, utilize an expectation-maximization technique to address label noise by leveraging over-parameterized models, though this method also incurs substantial additional computational costs. \textit{DivideMix}~\cite{li2020dividemix} and \textit{ProMix}~\cite{xiao2023promix} introduce techniques for probabilistic mixing of samples, aiming to filter noisy samples during training, but they still rely on computationally intensive procedures to maintain robust performance. The sample priors* framework~\cite{chen2023sample} employs sample reweighting based on prior distributions to discount noisy labels, but it similarly requires additional model components that limit its scalability.


The choice of batch size plays a crucial role in the trade-off between training stability and generalization. Studies by~\cite{mccandlish2018empirical} have shown that batch sizes up to a certain critical threshold stabilize model performance, whereas larger batch sizes tend to degrade generalization due to reduced gradient noise. Further studies have proposed batch-size scaling rules and scaling laws for adapting learning rate with batch size to optimize training efficiency and convergence~\cite{goyal2018accuratelargeminibatchsgd}.


Recent research has also focused on optimizing the gradient aggregation process itself. Techniques like gradient clipping~\cite{pascanu2013difficultytrainingrecurrentneural} help stabilize training by capping the norm of gradients, particularly in recurrent neural networks where gradient explosion is common. Further, gradient noise injection~\cite{neelakantan2015addinggradientnoiseimproves} has been explored as a means to escape sharp local minima and prevent overfitting. Our work builds on this line of inquiry by introducing gradient agreement filtering, a novel approach to dynamically filter micro-gradients based on cosine distance, allowing us to improve computational efficiency by reducing batch sizes while still maintaining training stability by excluding high-disagreement gradients in each macrobatch.

\section{Methods}
\label{sec:methods}

We consider the problem of how to most efficiently estimate an accurate gradient by aggregating micro-gradients during distributed training while preventing memorization and minimizing the compute budget. The core algorithm is presented in \Cref{alg:gaf}. Consider a training set $\mathcal{N}$ of size $n$. In traditional SGD, an update to the model parameters $\theta$ is computed by sampling a minibatch $\mathcal{B} \subset \mathcal{N}$ of size $|\mathcal{B}| = b$, calculating the gradient $\nabla_\theta \mathcal{L}(\mathcal{B}; \theta)$, and applying the following update rule
\begin{equation} \label{eq:sgd_update}
    \theta \leftarrow \theta - \eta \nabla_\theta \mathcal{L}(\mathcal{B}; \theta)
\end{equation}
where $\eta$ is the learning rate, and $\mathcal{L}(\mathcal{B}; \theta)$ is the loss function over the minibatch $\mathcal{B}$.

Due to GPU memory constraints, training is parallelized across multiple GPUs by computing the gradient for a \textit{macrobatch} of data comprised of multiple \textit{microbatches}. A microbatch $\mathcal{U}_i$ is a subset of samples within a larger macrobatch $\mathcal{M}$ where the microbatch data is small enough to fit in the VRAM of a single GPU. Each microbatch has size $|\mathcal{U}_i| = u$, a macrobatch $\mathcal{M}$ consists of multiple microbatches, i.e., $\mathcal{M} = \{\mathcal{U}_1, \mathcal{U}_2, \dots, \mathcal{U}_k\}$ with $|\mathcal{M}| = m = k \cdot u$. Typically $u \ll m$.

For each microbatch $\mathcal{U}_i$, a micro-gradient $\nabla_\theta \mathcal{L}(\mathcal{U}_i; \theta)$ is computed. The final gradient used to update $\theta$ is obtained by averaging the micro-gradients across all microbatches in $\mathcal{M}$
\begin{equation} \label{eq:macro_batch_eq_without_GAF}
    \nabla_\theta \mathcal{L}(\mathcal{M}; \theta) = \frac{1}{k} \sum_{i=1}^k \nabla_\theta \mathcal{L}(\mathcal{U}_i; \theta).
\end{equation}
The SGD update with the macrobatch gradient is then
\begin{equation} \label{eq:sgd_macrobatch_update}
    \theta \leftarrow \theta - \eta \nabla_\theta \mathcal{L}(\mathcal{M}; \theta).
\end{equation}

\begin{algorithm}[h]
\caption{Gradient Agreement Filtering (GAF)}
\label{alg:gaf}
\begin{algorithmic}[1]
\STATE \textbf{Input:} Training set $\mathcal{N}$, macrobatch size $m$, microbatch size $u$, training GPUs $k$, cosine distance threshold $\tau$, learning rate $\eta$, total training steps $T$
\FOR{$t \in [1, T]$}
    \STATE \textsc{sample} $\mathcal{M}_t \sim \mathcal{N}, \;$ {\upshape{s.t.}} $\; |\mathcal{M}_t| = m$
    \STATE \textsc{distribute} $\mathcal{M}_t $ \upshape{into} $k \; $\upshape{microbatches} $\mathcal{U}_k$
    \STATE \quad $\text{s.t.} \; \bigcup_{i=1}^{k} \mathcal{U}_{i} = \mathcal{M}_t, |\mathcal{U}_{i}| = u, m = u \times k$
    \STATE \textsc{sample} $s \sim \;$ \textsc{categorical}$(1, 2, \dots, k)$
    \STATE $\mathbf{g} \leftarrow \nabla_\theta \mathcal{L}(\mathcal{U}_s; \theta)$
    \STATE $\mathbf{c} \leftarrow 1$
    \FOR{$i \in [1, k], i \neq s$}
        \STATE $\mathbf{g}_i \leftarrow \nabla_\theta \mathcal{L}(\mathcal{U}_i; \theta)$
        \STATE $D_c(\mathbf{g}_i,\mathbf{g}) \leftarrow 1 - \frac{\mathbf{g}_i^T \mathbf{g}}{\|\mathbf{g}_i\| \|\mathbf{g}\|}$
        \IF{$D_c(\mathbf{g}_i,\mathbf{g}) \leq \tau$}
            \STATE $\mathbf{g} \leftarrow \mathbf{g} + \mathbf{g_i}$
            \STATE $\mathbf{c}   \leftarrow \mathbf{c} + 1$
        \ENDIF
        \IF{$c > 1$}
            \STATE $\mathbf{g}_{\text{GAF}} \leftarrow \frac{\mathbf{g}}{c}$
            \STATE $\theta \leftarrow \theta - \eta \mathbf{g}_{\text{GAF}}$
        \ELSE
            \STATE \textsc{continue}
        \ENDIF
    \ENDFOR
\ENDFOR
\end{algorithmic}
\end{algorithm}

\subsection{Gradient Agreement Filtering (GAF)}

\begin{figure}[th]
    \centering
    \begin{subfigure}{0.975\linewidth}
        \centering
        \includegraphics[width=\linewidth,  trim=0 0 0 17mm, clip]{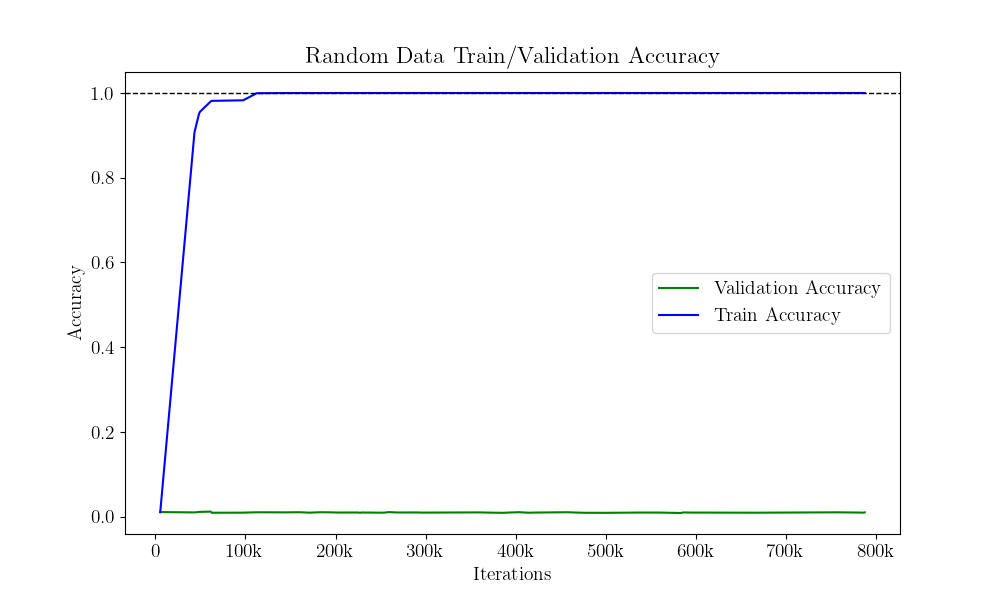}
    \end{subfigure}
    \hfill
    \begin{subfigure}{0.975\linewidth}
        \centering
        \includegraphics[width=\linewidth,  trim=0 0 0 17mm, clip]{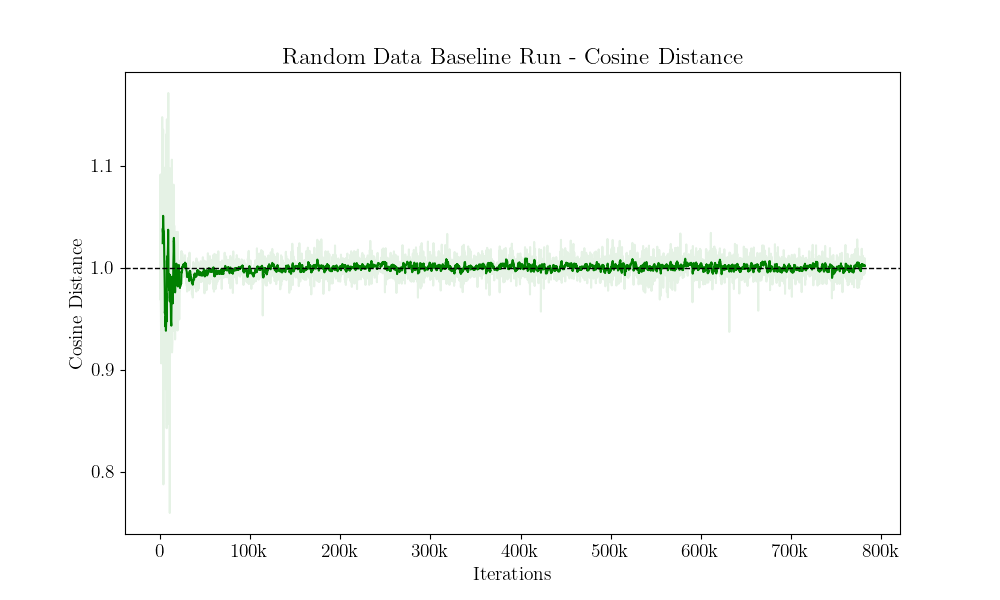}
    \end{subfigure}
   
    \caption{Train and validation accuracy (top) and the cosine distance between micro-gradients (bottom) with rolling average in dark green and raw values in light green, over iterations of a baseline training ResNet18 without GAF on random noise. The model overfits, reaching 100\% training accuracy, but the micro-gradients cosine distance remains above 0.96 throughout the entire training, and above 0.99 for all iterations after the very early iterations. }
    \label{fig:random_data_run}
\end{figure}

Gradient agreement filtering is an approach to aggregate micro-gradients that improves upon simply averaging all micro-gradients $\nabla_\theta \mathcal{L}(\mathcal{U}_i; \theta) \; \forall \; \mathcal{U}_i \in \mathcal{M}$. The approach is motivated by the following observation. If we train on completely random data (white noise), a model will overfit the train set but cosine distance will never fall below 0.99 after just a few iterations, as seen in \Cref{fig:random_data_run}. This suggests that we can prevent overfitting on noise by simply skipping updates where the micro-gradients have greater than 0.99 cosine distance. The cosine distance $D_c$ between two vectors $\mathbf{x}$ and $\mathbf{y}$ is
\begin{equation} \label{eq:cdist}
    D_c(\mathbf{x},\mathbf{y}) = 1 - \frac{\mathbf{x}^T\mathbf{y}}{\|\mathbf{x}\|\|\mathbf{y}\|}
\end{equation}

With Gradient Agreement Filtering (GAF), instead of blindly averaging all micro-gradients in $\mathcal{M}$, we apply a cosine distance threshold to select only those micro-gradients that are aligned within a given threshold $\tau$, as shown in \Cref{alg:gaf}. Let $\mathbf{g}_i = \nabla_\theta \mathcal{L}(\mathcal{U}_i; \theta)$ denote the micro-gradient for microbatch $\mathcal{U}_i$. The cosine distance between a candidate micro-gradient $\mathbf{g}_i$ and the running sum of accepted gradients $\mathbf{g}$ is $D_c(\mathbf{g}_i, \mathbf{g})$.

We compute a rolling aggregation of micro-gradients starting from the local gradient $\mathbf{g}$ and then checking one by one, and only including those for which $D_c(\mathbf{g}_i, \mathbf{g}) \leq \tau$. We keep a counter $c$ of the agreed upon gradients starting at $c = 1$. Each accepted gradient $\mathbf{g}_i$ is added to the running sum $\mathbf{g}$, and our count $c$ is incremented to keep track of the number of accepted gradients. The filtered macrobatch gradient is
\begin{equation} \label{eq:gaf_grad_update}
   \nabla_\theta \mathcal{L}_{\text{GAF}}(\mathcal{M}; \theta) = \mathbf{g}_{\text{GAF}} = \frac{\mathbf{g}}{c}.
\end{equation}

If no two gradients meet the threshold $\tau$ then $c = 1$ and we skip the update without modifying the optimizer or scheduler as we do not have consensus of any two micro-gradients. Otherwise, the GAF-based SGD update is
\begin{equation}
    \theta \leftarrow \theta - \eta \mathbf{g}_{\text{GAF}}.
\end{equation}

Note, that this implementation is order dependent so could be susceptible to degenerate examples. For example if the initial micro-gradient is orthogonal to all others, then none will agree and the entire macrobatch will be skipped and wasted. This is a shortcoming that could be addressed by tweaking the AllReduce algorithm such that each microgradient acts as the ``initial'' micro-gradient starting from its home GPU, and goes around the ring. The summed micro-gradient with the largest (or smallest) agreement could be the one that is then AllGather'd to the rest of the GPUs. We leave possible implementation to future research.

\section{Experiments}
\label{sec:experiments}

To demonstrate the effectiveness of GAF in practice, we train ResNet image classifiers on the CIFAR-100 and CIFAR-100N-Fine datasets using distributed data-parallelism comparing both baseline averaging-based gradient aggregation and GAF-based gradient aggregation. 

\begin{table*}[ht!]
\centering
\begin{tabular}{@{}lcccccccccc@{}}
\toprule
Dataset & \multicolumn{9}{c}{CIFAR-100} & CIFAR-100N-Fine \\
\midrule
Label Error & 0\% & 5\% & 15\% & 20\% & 30\% & 40\% & 50\% & 60\% & 75\% &  \\
\midrule
GAF & 63\% & 61\% & 60\% & 58\% & 54\% & 53\% & 48\% & 39\% & 13\% & 61.4\%  \\
Averaging & 62\% & 58\% & 53\% & 50\% & 40\% & 38\% & 33\% & 20\% & 11\% & 52.1\% \\
\bottomrule
Improvement & +0.2\% & +3\% & +7\% & +8\% & +14\% & +15\% & +15\% & +19\% & +2\% & +9.3\% \\
\bottomrule
\end{tabular}%
\caption{Image classification validation accuracy of ResNet18 on CIFAR-100 and CIFAR-100N-Fine when trained with GAF-based vs. averaging of micro-gradients. Improvement is the absolute increase in validation accuracy of GAF-based training over the baseline averaging.}
\end{table*}

\subsection{CIFAR-100}

We train RestNet18 on two A40 GPUs on the CIFAR-100 dataset using SGD with momentum and and reduction of the learning rate (learning rate) on validation plateaus with schedule patience of 100 and 0.1 discount. We use an initial learning rate of 0.01. We also applied L2 regularization with a weight decay of 0.01. In all cases, unless otherwise specified, we use a macrobatch of size $m = 200$ with $u = 100$ images per microbatch (exactly 1 sample per class) to ensure each microbatch has the same distribution of data over the training set. We flip each label with a random other classes label for x\% of the labels, for $ x \in \{0,5,15,30,40,50,60,75,80,90\}$, and maintain those incorrect label for the entirety of that run's training (i.e. symmetric noise). For each experiment we found the optimal value of the cosine distance threshold hyperparameter $\tau$ by performing a grid search of values from 0.95 to 1.05 with a step of 0.02 across different batch sizes. We compare with a baseline-training of ResNet18 where the cosine distance threshold is set to 2, which admits all gradients and is equivalent to averaging training weights when $k = 2$. We run for 500k iters for all runs, and observe convergence around 270k iterations for baseline and GAF runs. 

For the no error case, we observe cosine distance threshold of 1 yields best performance. Once errors are introduced we observe a cosine distance of 0.97 provides the best performance. 

\begin{figure}[t]
    \centering
    \begin{tikzpicture}
        \begin{axis}[
            width=0.9\linewidth,
            height=8cm,
            xlabel={Label Error (\%)},
            ylabel={Validation Accuracy (\%)},
            xmin=0, xmax=90,
            ymin=0, ymax=70,
            xtick={0, 10, 20, 30, 40, 50, 60, 70, 80, 90},
            ytick={0, 10, 20, 30, 40, 50, 60},
            yticklabel={\pgfmathprintnumber{\tick}\%},
            legend style={font=\small, at={(0.6, 0.95)}, anchor=north west},
            grid=major,
            label style={font=\small},
            tick label style={font=\small}
        ]

        \addplot[
            color=blue,
            mark=square,
            thick
        ] coordinates {
            (0, 63) (5, 61) (15, 60) (20, 58) (30, 54) (40, 53) (50, 48) (60, 39) (75, 13) (80, 8) (90, 4)
        };
        \addlegendentry{GAF}

        \addplot[
            color=red,
            mark=triangle,
            thick
        ] coordinates {
            (0, 62) (5, 58) (15, 53) (20, 50) (30, 40) (40, 38) (50, 33) (60, 20) (75, 11) (80, 7) (90, 1)
        };
        \addlegendentry{Baseline}

        \end{axis}
    \end{tikzpicture}
    \caption{Validation accuracy on CIFAR-100 with symmetric noisy labels for ResNet18 trained with and without GAF.}
    \label{fig:cifar100_over_noise_results}
\end{figure}
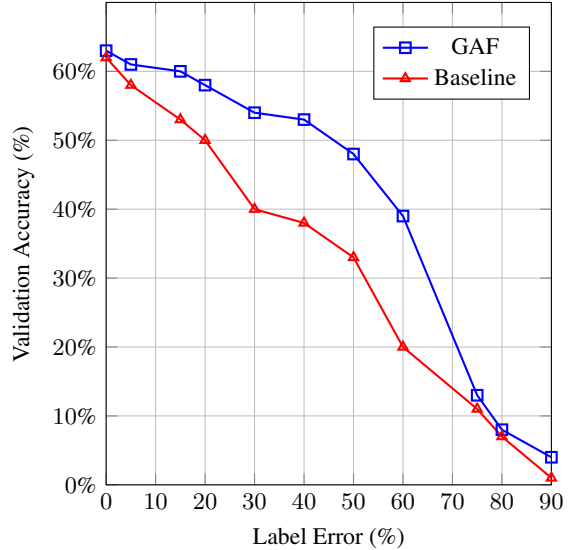

As shown in \Cref{fig:cifar100_over_noise_results}, we see a 0.2\% improvement over baseline without added noise to the CIFAR-100 dataset. As we add more and more noise to the labels, the GAF-trained models show increasing improvement over baseline until 60\% where it beats baseline model by 18.4\% when $\tau = 0.97$.

\begin{figure}[ht!]
    \centering
    \begin{tikzpicture}
        \begin{axis}[
            xlabel={Cosine Distance Threshold},
            ylabel={Validation Accuracy (\%)},
            xmin=0.94, xmax=1.06,
            ymin=50, ymax=65,
            xtick={0.95, 0.97, 0.99, 1.01, 1.03, 1.05},
            ytick={50, 55, 60, 65},
            legend pos=south west,
            grid=both,
            width=8cm,
            height=8cm,
        ]

        \addplot[
            color=blue,
            mark=*,
            thick,
        ] coordinates {
            (0.95, 60.76)
            (0.97, 60.91)
            (0.99, 58.99)
            (1.01, 58.27)
            (1.03, 58.19)
            (1.05, 58.02)
        };
        \addlegendentry{GAF (5\% Error)}

        \addplot[
            color=black,
            dashed,
        ] coordinates {
            (0.95, 58.57)
            (0.97, 58.57)
            (0.99, 58.57)
            (1.01, 58.57)
            (1.03, 58.57)
            (1.05, 58.57)
        };
        \addlegendentry{Baseline (5\% Error)}

        \addplot[
            color=red,
            dashed,
        ] coordinates {
            (0.95, 62.30)
            (0.97, 62.30)
            (0.99, 62.30)
            (1.01, 62.30)
            (1.03, 62.30)
            (1.05, 62.30)
        };
        \addlegendentry{Baseline (No Error)}

        \end{axis}
    \end{tikzpicture}
    \caption{Validation accuracy of ResNet18 runs trained on CIFAR-100 with GAF over different cosine distance thresholds. As the cosine distance threshold increases beyond 0.97 GAF-based training averages over more noise so generalization decreases. }
     \label{fig:cifar100_validation_over_cosine_distance}
\end{figure}
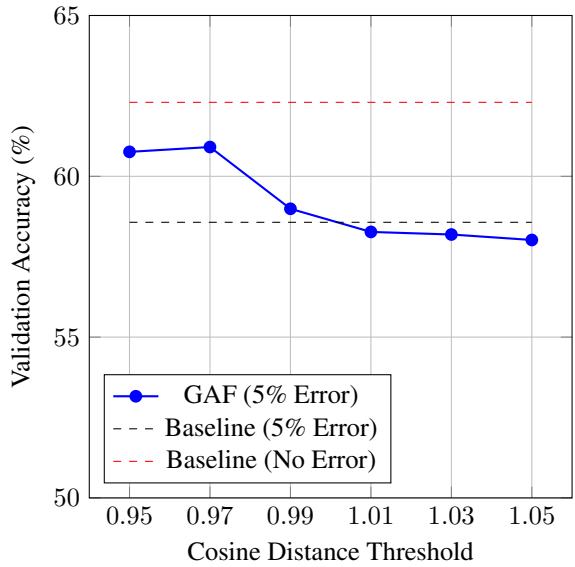

Additionally, we see in \Cref{fig:cifar100_validation_over_cosine_distance} that the performance improvement from GAF-based training ultimately decreases as we increase our cosine distance threshold. As we increase cosine distance threshold beyond 0.97, the improvement from GAF filtering goes away as the filter starts admitting more noise in our gradients, and removing the ability to discern good from bad micro-gradients.

\subsection{CIFAR-100N-Fine}

To validate GAF on a more realistic noisy dataset, we trained ResNet34 on CIFAR-100N-Fine. CIFAR-100N-Fine is a relabeled version with human annotated noisy labels obtained from one Amazon Mechanical Turk worker, who had a 40.2\% noise rate but in a more structured manner than random as humans are consistently biased in their labels vs. the random flipping done in the CIFAR-100 runs. All CIFAR-100N-Fine training runs use a ResNet34 with PreAct as per the reference paper \cite{wei2022learning}, trained on two A40 GPUs. We additionally test the effect of microbatch size $u$ on the training process by training with and without GAF for batch sizes of $u \in \{100, 200, 300, 400, 500\}$ As with the CIFAR-100 training, we use SGD with momentum and reduce the learning rate on validation plateaus. All other hyper parameters are the same as the CIFAR-100 runs however we do not vary label error percentage since the dataset is already noisy due to the labeling process. The optimal cosine distance threshold parameter $\tau$ is found by varying the value from 0.95 to 1.05 with a step of 0.02. A cosine distance threshold of 2 for the baseline runs, which is equivalent to averaging gradients as it admits all values.



\begin{figure}[ht!]
    \centering
    \begin{tikzpicture}
        \begin{axis}[
            xlabel={Microbatch Size (Samples per class per batch)},
            ylabel={Validation Accuracy (\%)},
            xmin=0, xmax=600,
            ymin=50, ymax=65,
            xtick={100, 200, 300, 400, 500, 600},
            ytick={50, 55, 60, 65},
            legend style={at={(0.975,0.975)}, anchor=north east, font=\tiny},
            grid=both,
            width=0.9\columnwidth,
            height=7cm,
        ]

        \addplot[
            color=blue,
            mark=*,
            thick,
        ] coordinates {
            (100, 60.79)
            (200, 58.37)
            (300, 55.36)
            (400, 52.95)
            (500, 51.63)
        };
        \addlegendentry{GAF}

        \addplot[
            color=red,
            mark=*,
            thick,
        ] coordinates {
            (100, 52.64)
            (200, 53.04)
            (300, 52.26)
            (400, 50.90)
            (500, 50.81)
        };
        \addlegendentry{Baseline}

        \addplot[
            color=black,
            dashed,
            mark=none,
            thick,
            nodes near coords={
                \pgfmathparse{\coordindex == 0 ? "0.97" : "0.95"} \pgfmathresult
            },
            every node near coord/.append style={font=\small, anchor=south, yshift=3pt}
        ] coordinates {
            (100, 61) 
            (200, 59)
            (300, 59)
            (400, 59)
            (500, 59)
        };
        \addlegendentry{Optimal Cosine Distance Threshold (Scaled)}

        \end{axis}
    \end{tikzpicture}
    
    \caption{ResNet34 PreAct Validation accuracy over microbatch size in the GAF and baseline case on CIFAR-100N-Fine, overlaid with cosine distance threshold used in training. As we increase microbatch size, the benefit of GAF reduces due to averaging more noisy samples.}
    \label{fig:cifarn-microbatch-size}
\end{figure}
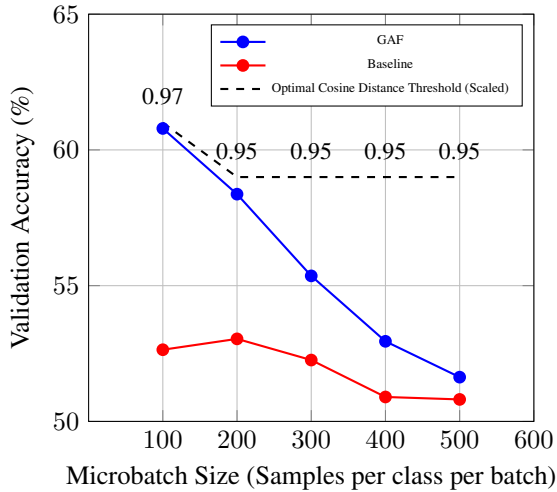

\Cref{fig:cifarn-microbatch-size} displays the result of these experiments. We find improvement in validation accuracy of training with GAF for all batch sizes, with the largest improvement in validation accuracy of 61.41\% with GAF vs. 52.1\% baseline accuracy for a microbatch size of $u = 100$, which provides a 9.3\% improvement. Note, this is at the smallest microbatch size possible that still contains at least one sample per class (100), while the best performing microbatch size for non-GAF was 200. This means we achieve higher accuracy with half the compute required and could have used half the GPUs (assuming we used multiple processes per GPU). 

\begin{figure}[h]
    \centering
    \includegraphics[width=0.975\linewidth]{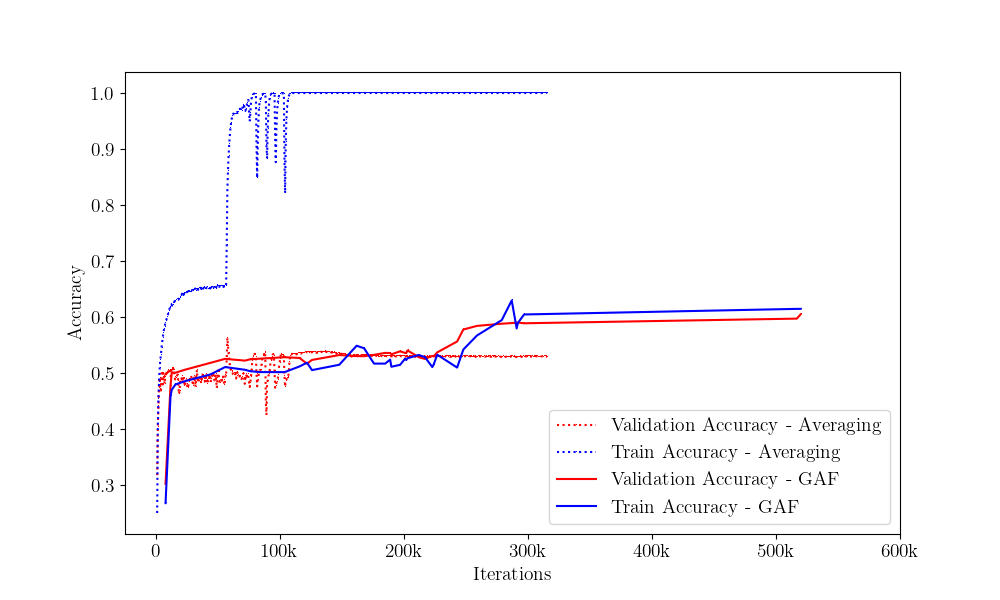}
    \caption{Train (blue) and validation accuracy (red) over both GAF (solid) and averaging (dotted) runs of ResNet34 of CIFAR-100N-Fine. GAF train accuracy remains very close to validation accuracy, while averaging results in overfitting to the train set within 100k iterations. GAF continues to improve on validation after 500k iterations, albeit very slowly.}
    \label{fig:cifarn_does_not_overfit_plot}
\end{figure}

Additionally, baseline training on CIFAR-100-N-Fine plateaus at 52.1\% validation accuracy with 100\% train accuracy at 150k iterations. However, even after 600k iterations, GAF-based training surprisingly does not overfit with 59.6\% train accuracy 61.4\% validation accuracy, with slow but continued improvement in both, as shown in \Cref{fig:cifarn_does_not_overfit_plot}.

Finally, we find that the performance improvement from GAF degrades as we increase microbatch size. This means that training can be done with smaller batch sizes and that larger batch sizes in training only increases computational costs without benefit. As with the CIFAR-100 training experiments, the higher we make microbatch size, the benefit of GAF decreases as we begin averaging over more noise and removing the ability for GAF to discern good from bad microbatches. Consequently, we also find that the optimal cosine distance threshold decreases from 0.97 to 0.95 as batch size increases to further increase the filtering as the micro-gradients become increasing correlated. Thus when doing training with GAF following the typical procedure of choosing the largest microbatch size that can fit on a single GPU results in lower validation and instead we should use smaller microbatch sizes and fewer GPUs to achieve higher levels of generalization.

This experiment shows that in addition to a 9.3\% improvement over baseline with a microbatch size of only 100, GAF-based training with smaller microbatches outperforms higher microbatch sizes enabling us to achieve improved training performance with an order of magnitude less compute. 

\section{Conclusions}
\label{sec:conclusions}

In this work, we introduced Gradient Agreement Filtering (GAF) as an alternative to traditional micro-gradient averaging in distributed training. Our experiments on CIFAR-100 and CIFAR-100N-Fine demonstrate the effectiveness of GAF, particularly in scenarios with label noise. By aggregating gradients based on cosine distance, GAF provides a robust approach that improves model performance. Specifically, we observe a 0.2\% improvement on CIFAR-100 without added noise, with progressively larger improvements over baseline training methods as label noise increases, reaching up to an 18.4\% gain at a 60\% noise rate. On the CIFAR-100N-Fine dataset, GAF achieves a 9.3\% improvement over the baseline. We also observe that we are able to maintain the performance improvement even as the microbatch size was reduced, suggesting that we can improve model performance while reducing computational costs.

These results indicate that GAF is a promising approach for improving training stability and accuracy, particularly in noisy environments. The use of cosine distance to filter gradients shows potential not only in mitigating the impact of label noise but also in reducing the computational cost of large-scale distributed training by focusing resources on more aligned gradients.

\section{Future Research Directions}
\label{sec:future_work}

While GAF has demonstrated promising results, several avenues for further research could expand upon its potential and applicability:

\begin{itemize}
    \item \textbf{Alternative Similarity Metrics}: While cosine distance proved effective, other similarity metrics, such as Mahalanobis distance, could be explored to evaluate their impact on GAF’s performance. This could help in tailoring GAF to different types of datasets and noise structures.

    \item \textbf{Adaptive Thresholding}: In this work, we used a fixed cosine distance threshold throughout training. An adaptive threshold that dynamically adjusts based on training progress or model convergence rates may yield improved results, especially in tasks with fluctuating noise levels or diverse data distributions.


    \item \textbf{Application to Other Tasks}: GAF was applied to image classification in this study. Extending this technique to other domains, such as natural language processing, speech recognition, or reinforcement learning, could uncover broader benefits and challenges associated with GAF in non-vision tasks. 

    \item \textbf{Memory and Computation Efficiency}: As GAF requires tracking only pairwise cosine distances between micro-gradients, applying this to Ring-AllReduce would be straightforward but would require applying cosine distance to buckets at a time. Ensuring GAF's improvement is maintained despite this is an area of future research, as well as other avenues to optimize compute and memory overhead.
    
    \item \textbf{Order Indifference Techniques}: As GAF is sensitive to the order in which microgradients are processed, perhaps there is a way to augment Ring-AllReduce where during the AllGather phase, the GPU with the highest (or lowest) agreement is the one distributed to all other nodes. 

    \item \textbf{Integration with Advanced Optimizers}: We used standard optimizers like SGD and Adam in our experiments. Investigating how GAF interacts with other advanced optimization techniques, such as Adam, AdamW, LAMB or SHAMPOO, could enhance GAF’s performance, particularly in large-scale or fine-tuning scenarios.

    \item \textbf{Analysis of Gradient Disagreement Dynamics}: Further study of the dynamics of gradient disagreement over the course of training could yield insights into how models converge under noisy conditions and how GAF influences the loss landscape. This might lead to improvements in convergence rates and generalization.

\end{itemize}

Further research in these directions highlight potential improvements and adaptations of GAF, aiming to make it more efficient, robust, and applicable across various deep learning domains. 

\section*{Acknowledgments}

We would like to acknowledge and thank Alex Tzikas, Harrison Delecki, and Francois Chollet who provided invaluable help through discussions and feedback.

{
    \small
    \bibliographystyle{ieeenat_fullname}
    \bibliography{main}
}

\clearpage 
\appendix 


\end{document}